# Distributed Bayesian inference for consistent labeling of tracked objects in non-overlapping camera networks

**Wan Jiuqing**

Department of Automation

Beijing University of Aeronautics and Astronautics

wanjiuqing@gmail.com

**Liu Li**

Department of Automation

Beijing University of Aeronautics and Astronautics

**Abstract**: One of the fundamental requirements for visual surveillance using non-overlapping camera networks is the correct labeling of tracked objects on each camera in a consistent way, in the sense that the captured tracklets, or observations in this paper, of the same object at different cameras should be assigned with the same label. In this paper, we formulate this task as a Bayesian inference problem and propose a distributed inference framework in which the posterior distribution of labeling variable corresponding to each observation, conditioned on all history appearance and spatio-temporal evidence made in the whole networks, is calculated based solely on local information processing on each camera and mutual information exchanging between neighboring cameras. In our framework, the number of objects presenting in the monitored region, i.e. the sampling space of labeling variables, does not need to be specified beforehand. Instead, it can be determined automatically on the fly. In addition, we make no assumption about the appearance distribution of a single object, but use "similarity" scores between appearance pairs, given by advanced object re-identification algorithm, as appearance likelihood for inference. This feature makes our method very flexible and competitive when observing condition undergoes large changes across camera views. To cope with the problem of missing detection, which is critical for distributed inference, we consider an enlarged neighborhood of each camera during inference and use a mixture model to describe the higher order spatio-temporal constraints. The robustness of the algorithm against missing detection is improved at the cost of slightly increased computation and communication burden at each camera node. Finally, we demonstrate the effectiveness of our method through experiments on an indoor Office Building dataset and an outdoor Campus Garden dataset.

# Distributed Bayesian inference for consistent labeling of tracked objects in non-overlapping camera networks

1. Introduction

Recently, there has been increasing research interest in wide-area video surveillance based on smart camera networks with non-overlapping Field of View (FOV). The cameras in the networks are not only able to capture video data, but also capable of local processing and mutual communication. They usually work cooperatively for discovering and understanding the behavior of some interested objects, e.g. pedestrians, vehicles, in the monitored region. One of the fundamental prerequisite for achieving this goal is the correct labeling of the observations of objects captured by each camera nodes, in a consistent way. That is, the observations assigned with the same label are assumed to be generated form the same object. In this paper, we assume that the detection and tracking problem within a single camera view has been already solved, and we call some interested quantities extracted from the tracked object as an virtual "observation", see Fig.2.

Consistent labeling of tracked objects in non-overlapping camera networks is, however, a rather challenging task. First, the object's appearance often undergoes large variations across disjoint camera views due to significant changes in view angle, lighting, background clutter and occlusion. The different objects may appear more alike than that of the same object across different views. This makes the labeling of objects based solely on appearance cues to be very difficult. Second, for large scale application, it is unrealistic to transmit all video data collected by cameras in the networks to a central server for labeling inference due to the limitation of communication bandwidth and camera node energy. Even through smart camera can analysis the local video and transmit only extracted features, the central server will become overwhelmed quickly when the number of objects and measurements increases because of the combinatorial nature of labeling problem. Thus, distributed algorithms are preferred rather than centralized one. Third, the uncertainty in the number of objects presenting in the monitored region makes the labeling problem even more difficult, as we should infer not only the label for each tracked object but also the number of labels, i.e. how many objects are moving in the region, at the same time.

1.1 Related works

A lot of works have been proposed to answer the above challenges. In community of object re-identification, huge amounts of efforts [1-6] are devoted to matching objects across different views based on the unreliable appearance measurements. However, the results are still far from satisfactory and the outputs of re-identification algorithms are always in the form of a list of top ranked candidates, which can not be used directly for labeling. Recently, some works have been shown [7] in which spatio-temporal cues are exploited to improve matching accuracy. But in [7] the authors only consider matching problem between a pair of views, and extending it to camera networks is not a trivial matter.

The problem of consistent labeling in camera networks using both appearance and spatio-temporal cues have been widely investigated, under the name of data association [8-9], trajectory recovery [10], or camera-to-camera tracking [11-16]. Some authors try to solve the problem by optimally partitioning the set of observations collected by the camera networks into several disjoint subsets, such that the observations in each subset are believed to come from a single object. The difficulty of exponentially growth of the partition space is tackled by making appropriate independence assumptions and leveraging efficient optimization algorithms, such as Markov Chain Monte Carlo [10, 13], Max-Flow Network [14], or Multiple Hypothesis Tracking [15-16]. On the other hand, it is more attractive to treat the problem in a Bayesian framework [8-9], in which the posterior distribution of the labeling variable corresponding to each observation is inferred conditioned on all evidence made in the whole networks, as the resulting marginal distribution of labeling variable contains the complete knowledge about which object the observation is originated from. However, doing inference in the joint labeling space is usually intractable and the authors have to resort to some assumed independence structure and approximate algorithms such as Assumed Density Filter [8-9].

The above approaches are all centralized, making them not suitable for large scale networks for reasons mentioned before. Recently, distributed solutions of consistent labeling problem, which involves only local information processing and exchanging while being able to achieve the same or similar labeling accuracy as their centralized counterparts, have attracted many research interests. Considering the appearance observations made in the networks as i.i.d. samples drawn from a mixture model, and treating the labeling variables as missing data, various appearance-based distributed labeling methods have been proposed under different kinds of distributed EM frameworks [17-20]. However, these algorithms always perform poorly when observing conditions vary largely across camera views, as they assume that the appearance of a single object follows a unimodal distribution, corresponding to a component in the mixture model. To improve the labeling performance, exploiting the spatio-temporal information is necessary. Unfortunately, unlike the case of ordinary wireless sensor networks [21-27] or camera networks with overlapping FOVs [28-30], where the dependence of involved variables in spatial dimension (intra-scan dependence) and temporal dimension (intra-scan dependence) can be modeled separately, the spatial and temporal evidence made in non-overlapping camera networks are tightly coupled, which precludes the use of most existing distributed inference or optimization algorithms for WSN and overlapping camera networks. In our recent works [31], based on the non-missing detection assumption, we use a spatio-temporal tree to model the dependence of involved variables, and use belief propagation algorithm for calculating the posterior probability of labeling variable, i.e. observation ownership in E-step of the distributed EM framework. Compared with traditional distributed EM, significant performance gain has been obtained by the effective using of spatio-temporal information.

1.2 Our contributions

The main limitations of [31] are: (1) the number of objects under tracking needs to be known beforehand, and (2) the appearance of a single object is assumed to be unimodal distributed. In this paper, we propose a new distributed Bayesian inference framework for consistent labeling of the tracked objects in non-overlapping camera networks, which nicely overcomes the above limitations.

In our method, under the same non-missing detection assumption as in [31], the posterior distribution of the labeling variable conditioned on all of the appearance and spatio-temporal measurements made in the networks is calculated, based solely on local inference on each camera nodes and belief propagation between neighboring cameras. What makes this possible is that when the label of an observation made on a specific camera is inferred, all relevant information generated by the networks have been summarized in the belief states of labeling variables already located on the camera's neighbors.

Unlike [31], we do not prespecify a fixed number of possible objects, i.e., a fixed sampling space for each labeling variables. Instead, we allow each observation to ignite a new possible object, or equivalently, to add a new element in the sampling space, as we notice that each observation is either generated from a previously observed object, or from a newly appeared one. Based on the non-missing detection assumption mentioned before, the sampling space of current labeling variable is determined in an online and distributed manner, by combining those of labeling variables already located on neighboring cameras. Through the propagation of sampling spaces, it can be shown that each camera always perform inference in a space consisting of identifiers of all possible objects that may produce the current observation.

In addition, in this work we discard the unimodal assumption of object appearance used in [31]. Instead, we only assume that two observations are "similar" if they originate from the same object. Similar in appearance implies a higher rank in the output list of some object re-identification algorithm. Consequently, two observations of the same object that look quite different in original color space due to variance in illuminating conditions may get higher similarity score by leveraging advanced techniques developed in object re-identification community. Spatio-temporal similarity is determined by the level of fitting to the spatio-temporal model learned from training data or prespecified according to the prior knowledge of monitored region. In our Bayesian frameworks, the likelihood of observation is defined in terms of above similarity measures between observation pairs, through which the information made in the whole camera networks is injected elegantly into the process of labeling inference.

Conceptually, the sampling space of labeling variables will grow unlimited with the accumulation of observations, which may prevent the algorithm from being used in large scale applications. However, it is notable that the time separation between two successive observations of the same object cannot be arbitrarily large, and the number of objects is much smaller than that of observations. Accordingly, we set a limit of memory depth for each camera node by discarding the oldest observation, and control the size of sampling space of labeling variable by deleting elements with negligible posterior probability. In this way, we obtain an inference algorithm with constant computational and memory requirement, at very little cost of labeling accuracy.

Although many excellent works exist for object detection and tracking in single view, reliable detection in crowded scene is still a challenging task. Sometimes missing detection may happen and the assumption underlying our method is violated. We alleviate this problem by considering an enlarged neighborhood, and assuming that object has been detected at least once in this neighborhood before it arrives the current camera. We use a mixture model to describe the uncertainty in object moving path caused by observation missing and modify the evaluation of spatio-temporal likelihood, leading to improved robustness of the algorithm against missing detection.

Extensive experiments are conducted on two datasets collected by our non-overlapping camera networks: the Office Building dataset and Campus Garden dataset, and comparisons are made with two closed related inference-based consistent labeling algorithms. The results demonstrate that: (1) compared with centralized inference algorithm [8], our method shows significant superiority in execution speed, and achieves comparable labeling accuracy; (2) compared with distributed inference algorithm [31], our method provides the ability to estimate the number of moving objects and also shows obvious improvement in labeling accuracy; (3) by considering the higher order neighborhood, our method gives satisfactory results in case of missing detections.

2. Problem formulation

Suppose multiple objects are moving in a large area monitored by $N$ smart cameras with non-overlapping field of views, as shown in Fig.1. Here the number of moving objects is not fixed in time and unknown to us. We assume that each camera node has limited resource for computation, storage and communication, and synchronized internal clocks that allow the nodes to share a common notion of time. The camera networks can be represented as a graph $G = (V, E)$, which is called camera networks topology or activity topology, see Fig.1. Each camera corresponds to a node $v \in V$ in the graph, and two nodes are linked by an edge $e \in E$ if object can move from one node to another without being observed by other cameras. In this paper, camera node $v$ is called (zero-order) neighbor of node $u$ if $(u,v) \in E$. And we use $\mathcal{N}_u$ to denote the set of neighbors of $u$. The topological assumption imposes strong constraints on the movements of objects in the networks, in that any object, if not newly appeared, must presented in the FOV of one of current camera's neighbors before it arrives the current camera. It holds true in many scenarios of interest and has been adopted in most of the works about tracking in non-overlapping camera networks [8-16, 31].

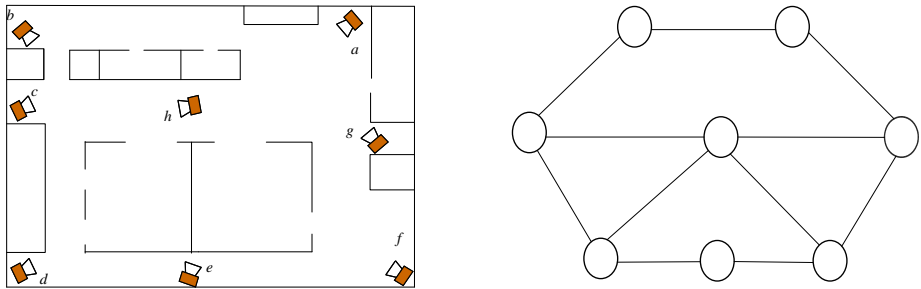

**Fig.1 Smart camera networks and its topology**

A clip of video is collected by camera when an object is passing through its FOV. We assume that the collected video clip has been summarized into a single virtual observation $y_{u,i}^k$, by some visual tracking and feature extraction algorithms running on the camera, such as [33,34]. For convenience of discussion, here the observation $y_{u,i}^k$ is double indexed: the local index $i$ implies that it is the $i$ th observation generated by camera $u$, and the global index $k$ implies that it is also the $k$ th observation generated in the whole networks. It should be noticed that each camera is aware of the local index of the observations made on itself, but not the global index. Each observation $y_{u,i}^k$ consists of two parts: the appearance measurement $o_{u,i}^k$, such as the size, texture, color distribution, or biometric features of the object; and the spatio-temporal measurement $d_{u,i}^k$, such as the capturing time or the moving direction of the object in the camera's FOV. An example is shown in Fig.2. After a certain period of time, a set of observations are generated on each camera in the networks. We use $Y^{1:k}$ to denote the observation set generated in the whole networks up to step $k$.

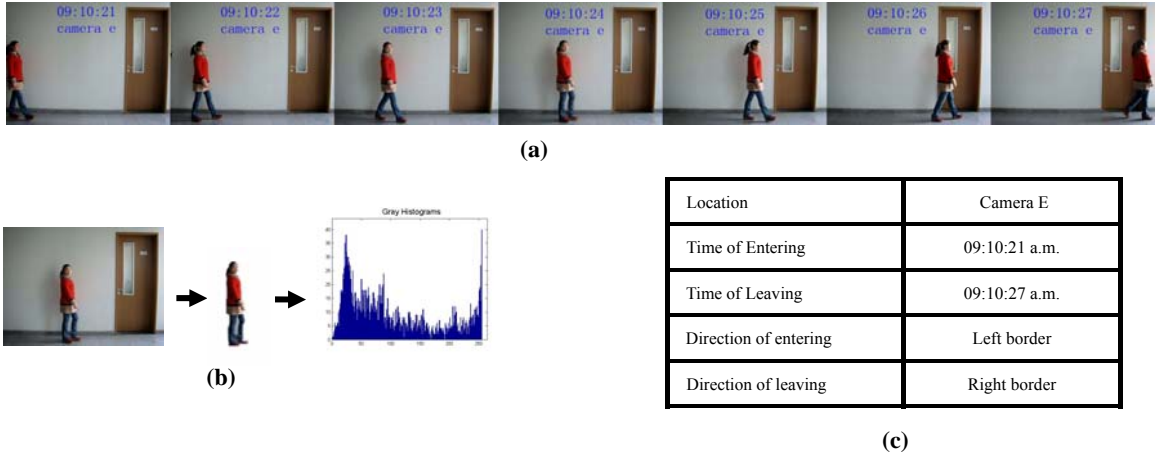

**Fig.2 Observations made by camera node. (a) Video frames collected by camera when an object is passing by. (b) Appearance observation: the color histogram of the object region segmented from frames. (c) Spatio-temporal observation: the time and direction of the object's entering in or leaving from the camera's FOV.**

For each observation $y_{u,i}^k$, we assign a random labeling variable $x_{u,i}^k$ to indicate which object the observation is coming from. The sampling space of $x_{u,i}^k$ consists of the identifiers, or labels (in this paper, the terms identifier and label are used interchangeably), of all possible objects that may produce observation $y_{u,i}^k$. And the labeling on different cameras should be consistent with each other, i.e., observations generated by a single object on different cameras should be assigned with the same label. Our goal is to infer the posterior distribution of each labeling variables $x_{u,i}^k$, based on observations made in the whole networks up to current time step, that is, $p(x_{u,i}^k | Y^{1:k})$. The label of each observation is determined by choosing the label value with the maximum posterior probability. And the camera-to-camera trajectory of each object can be recovered by associating the observations with the same label. Here the main difficulties are: (1) We cannot transmit all observations $Y^{1:k}$ to a central server for calculating $p(x_{u,i}^k | Y^{1:k})$. Instead, the calculation must be based on local processing on individual cameras and communication between neighboring ones. (2) As the number of moving objects is unknown *a prior*, the sampling space of labeling variables on each cameras should be determined and maintained consistent with each other on the fly and also in a distributed manner.

3. Inference algorithm

In this section, we present our distributed Bayesian inference framework for consistent labeling. We show how to determine the sampling space of each labeling variables in Sec.3.1 and how to perform inference in that space in Sec.3.2. We present a distributed online algorithm with constant computation and memory requirements in Sec.3.3, by limiting the memory depth and maximum number of objects. Finally, we discuss the problem of missing detection and alleviate it by enlarging the camera's neighborhood.

3.1 The sampling space

We denote the sampling space of labeling variable $x$ as $\Gamma(x)$. The elements in $\Gamma(x)$ are globally unique identifiers of all possible objects generating the corresponding observation $y$. We assume that each observation is generated by single object on a specific camera. If multiple objects present in camera' FOV simultaneously, we can tracked them with multi-object tracking algorithm and produce one observation for each object. Thus, the pair of camera and local time index, $(u,i)$, can be used as globally unique identifiers of objects generating observations. If an observation $y$ is labeled as $(u,i)$, this means that $y$ is originated from an object whose first observation in the networks is $y_{u,i}^k$. In other words, we use the head of the trajectory of the object as its identifier. As the pair $(u,i)$ is one-to-one corresponding to the global time index $k$, in the following discussion we use $k$ in place of $(u,i)$ as element of $\Gamma$ just for simplicity in expression.

Suppose at time step $k$, an observation $y_{u,i}^k$ is generated on camera $u$, and the corresponding labeling variable is $x_{u,i}^k$. Before $k$, a set of observations has been made on $u$ and $u$'s neighbors. We denote this set as $Y_{\mathcal{N}_u}^{k-1}$, and the corresponding labeling variable set as $X_{\mathcal{N}_u}^{k-1}$. If no missing detection occurs, the observation $y_{u,i}^k$ either originates from object that has been detected in $u$'s neighborhood, or from a newly appeared one. Consequently, the sampling space of $x_{u,i}^k$ should be $k$ plus the union of the sampling space of labeling variables in the set $X_{\mathcal{N}_u}^{k-1}$, that is

$$\Gamma\left(x_{u,i}^k\right) = \{k\} \cup \left(\bigcup_{l=1:L} \Gamma\left(X_{\mathcal{N}_u,l}^{k-1}\right)\right) \tag{1}$$

where $X_{\mathcal{N}_u,l}^{k-1}$ represents the $l$ th element in the set $X_{\mathcal{N}_u}^{k-1}$. We can see that by using Eq.(1) the sampling space of $x_{u,i}^k$ can be determined automatically based on local information on $u$ and its neighborhood. It is easy to verify that if no missing detection occurs, each camera node is ensured performing inference in a sampling space consisting of identifiers of all possible objects generating the current observation.

3.2 The posterior

In this subsection we will discuss how to calculate the belief state, i.e., the posterior distribution of the labeling variable over its sampling space, conditioned on all observations made in the whole networks up to the current time step. To clarify the inter-observation dependence, for each $x_{u,i}^k$ we introduce a pointer variable $z_{u,i}^k$ to indicate the immediate predecessor of observation $y_{u,i}^k$ in the trajectory of a single object. The sampling space of $z_{u,i}^k$ is $\{0,1,\ldots,L\}$, $L$ is the cardinality of the set $Y_{\mathcal{N}_u}^{k-1}$. $z_{u,i}^k = 0$ means $y_{u,i}^k$ is the first observation of a newly appeared object. $z_{u,i}^k = l \neq 0$ means that the last observation of the same object directly before $y_{u,i}^k$ is the $l$ th element in the set $Y_{\mathcal{N}_u}^{k-1}$. In this paper we assume that the elements in $Y_{\mathcal{N}_u}^{k-1}$ and $X_{\mathcal{N}_u}^{k-1}$ have been sorted in reversed time order.

Using the Bayes rule, the joint belief state of $x_{u,i}^k$ and $z_{u,i}^k$ can be written as

$$\begin{aligned} b\left(x_{u,i}^k = h, z_{u,i}^k = l\right) &\triangleq p\left(x_{u,i}^k = h, z_{u,i}^k = l \mid Y^{1:k}\right) \\ &\propto p\left(y_{u,i}^k \mid x_{u,i}^k = h, z_{u,i}^k = l, Y^{1:k-1}\right) p\left(x_{u,i}^k = h, z_{u,i}^k = l \mid Y^{1:k-1}\right) \\ &= \lambda_{ap}\left(o_{u,i}^k, o_{u',i'}^{k'}\right) \lambda_{st}\left(d_{u,i}^k, d_{u',i'}^{k'}\right) p_r\left(x_{u,i}^k = h, z_{u,i}^k = l\right) \end{aligned} \tag{2}$$

here we assume that that the $l$ th observation in $Y_{\mathcal{N}_u}^{k-1}$ is $y_{u',i'}^{k'}$, and the appearance and spatio-temporal parts of the observation are conditionally independent given the hidden states. The term $\lambda_{ap}$ is the likelihood of appearance observation given $z_{u,i}^k = l$,

$$\lambda_{ap}\left(o_{u,i}^k, o_{u',i'}^{k'}\right) \triangleq p\left(o_{u,i}^k \mid z_{u,i}^k = l, Y^{1:k-1}\right) \tag{3}$$

it is a similarity measure of the pair $o_{u,i}^k$ and $o_{u',i'}^{k'}$. In this paper, we take the normalized histograms of RGB brightness values within the object region as appearance observation and use the bi-directional Cumulative Brightness Transfer Functions (CBTF) proposed in [1] for calculating the appearance likelihood. By establishing a mapping of brightness values between camera views, the CBTF compensates the variation in illumination conditions at different camera sites. Here we use CBTF for evaluating $\lambda_{ap}$ for its simplicity and effectiveness. In fact, any approach for the problem of appearance based people re-identification across disjoint cameras[1-6], which is a interesting topic attracting intensive research efforts recently, can serve for this purpose as long as they can output a similarity measure between a pair of appearance observations.

The term $\lambda_{st}$ is the spatio-temporal likelihood given $z_{u,i}^k = l$. Supposing that the object entered the FOV of camera $u$ at time $t_{u,i}^{en}$ via frame border $e_{u,i}^{en}$, and leaved it at time $t_{u,i}^{le}$ via border $e_{u,i}^{le}$. The spatio-temporal likelihood can be written as

$$\lambda_{st}\left(d_{u,i}^k, d_{u',i'}^{k'}\right) \triangleq p\left(d_{u,i}^k \mid z_{u,i}^k = l, Y^{1:k-1}\right) \\ = p\left(t_{u,i}^{en} \mid t_{u',i'}^{le}\right) p\left(e_{u,i}^{en} \mid e_{u',i'}^{le}\right) \quad (4)$$

Here $d_{u',i;}^{k'} = \left(t_{u',i'}^{en}, e_{u',i'}^{en}, t_{u',i'}^{le}, e_{u',i'}^{le}\right)$. The first factor $p\left(t_{u,i}^{en} \mid t_{u',i'}^{le}\right)$ in Eq.(4) models the time of arriving at camera $u$ knowing that an object left camera $u'$ at time $t_{u',i'}^{le}$,

$$p\left(t_{u,i}^{en} \mid t_{u',i'}^{le}\right) = \begin{cases} 0, & \text{if } t_{u,i}^{en} \leq t_{u',i'}^{le} + \Delta_{u,u'} \\ N\left(t_{u,i}^{en} - t_{u',i'}^{le} \mid \delta_{u,u'}, R_{u,u'}\right), \text{otherwise} \end{cases} \quad (5)$$

where $\Delta_{u,u'}$ is the minimum travel time between $u$ and $u'$, $\delta_{u,u'}$ and $R_{u,u'}$ are the expected travel time and the variance of this distribution. Here a truncated Gaussian distribution is used to prevent unrealistic travel time between cameras. The second factor $p\left(e_{u,i}^{en} \mid e_{u',i'}^{le}\right)$ in Eq.(4) is a discrete distribution specifying the probability of an object arriving at camera $u$ via border $e_{u,i}^{en}$ when departing from camera $u'$ via border $e_{u',i'}^{le}$. The entries of this distribution and the parameters in Eq.(5), $\left(\Delta_{u,u'}, \delta_{u,u'}, R_{u,u'}\right)$, are specified according to the prior knowledge of the layout of the region monitored by the camera networks, or learned form training data. When $z_{u,i}^k = 0$, there is no observation before $y_{u,i}^k$ in the same trajectory, the likelihood in Eq.(2) takes a constant value of $\lambda_0$, which is determined experimentally.

The prior in Eq.(2) can be factorized as follows

$$p_r\left(x_{u,i}^k = h, z_{u,i}^k = l\right) = p_r\left(x_{u,i}^k = h\right) p_r\left(z_{u,i}^k = l \mid x_{u,i}^k = h\right) \quad (6)$$

The term $p_r\left(x_{u,i}^k = h\right)$ is the predictive distribution of $x_{u,i}^k$ before observation $y_{u,i}^k$ is made. From the non-missing detection assumption mentioned before, $x_{u,i}^k$ should either take value of $k$ if $y_{u,i}^k$ is from a newly appeared object, or take the same value as one element in $X_{\mathcal{N}_u}^{k-1}$, if $y_{u,i}^k$ is from an older object. In the latter case, to predict the label of current observation, we first pick a variable $X_{\mathcal{N}_u,l}^{k-1}$ randomly from the set $X_{\mathcal{N}_u}^{k-1}$, then draw a labeling value according to its belief state, $p\left(X_{\mathcal{N}_u,l}^{k-1} \mid Y^{1:k-1}\right)$, which has been calculated in previous inference step. Consequently, the prior $p_r\left(x_{u,i}^k = h\right)$ can be calculated as

$$p_r\left(x_{u,i}^k = h\right) = \begin{cases} \dfrac{1}{L+1}, & h = k \\ \dfrac{1}{L+1} \sum_{l=1}^{L} p\left(X_{\mathcal{N}_u,l}^{k-1} = h \mid Y^{1:k-1}\right), & h \in \bigcup_{l=1:L} \Gamma\left(X_{\mathcal{N}_u,l}^{k-1}\right) \end{cases} \quad (7)$$

The term $p_r\left(z_{u,i}^k = l \mid x_{u,i}^k = h\right)$ is the predictive distribution of pointer variable $z_{u,i}^k$ knowing that the label of current observation is $h$. From the definition of $z_{u,i}^k$, it is easy to see that given $X_{\mathcal{N}_u}^{k-1}$, $z_{u,i}^k$ is independent of the information outside camera $u$'s neighborhood. Specifically, the prior $p_r\left(z_{u,i}^k = l \mid x_{u,i}^k = h\right)$ can be written as

$$p_r\left(z_{u,i}^k = l \mid x_{u,i}^k = h\right) = \begin{cases} p\left(X_{\mathcal{N}_u,1:L}^{k-1} \neq h \mid Y^{1:k-1}\right), & l = 0 \\ p\left(X_{\mathcal{N}_u,1:l-1}^{k-1} \neq h, X_{\mathcal{N}_u,l}^{k-1} = h \mid Y^{1:k-1}\right), & l = 1, \ldots, L \end{cases} \quad (8)$$

In Eq.(8) we need to evaluate the joint distribution of elements in the set $X_{\mathcal{N}_u}^{k-1}$, which is intractable when $L$ is large. Thus we approximate it using the product of marginal distributions of individual labeling variables

$$p\left(X_{\mathcal{N}_u}^{k-1}\middle|Y^{1:k-1}\right) \approx \prod_l p\left(X_{\mathcal{N}_u,l}^{k-1}\middle|Y^{1:k-1}\right) \tag{9}$$

The approximation scheme in Eq.(9) is similar with that of the Assumed Density Filter used in [8,9], in that the joint belief of hidden variables is repeatedly factorized into an assumed density family, i.e., the product of marginals, in each inference step. It is clear in the above discussion that, the likelihood and prior can be calculated based solely on local information, i.e., the observation made on camera $u$ and its neighbors and the beliefs of the corresponding labeling variables.

By summarizing out the auxiliary variable $z_{u,i}^k$ from Eq.(2), we get the belief of labels

$$b\left(x_{u,i}^k = h\right) \triangleq p\left(x_{u,i}^k = h\middle|Y^{1:k}\right)$$
$$\propto p_r\left(x_{u,i}^k = h\right) \sum_{l=0}^{L} \underbrace{\underbrace{\lambda_{ap}\left(o_{u,i}^k, o_{u',i'}^{k'}\right) \lambda_{st}\left(d_{u,i}^k, d_{u',i'}^{k'}\right)}_{\lambda\left(y_{u,i}^k, y_{u',i'}^{k'}\right) \triangleq p\left(y_{u,i}^k\middle|z_{u,i}^k = l\right)} p_r\left(z_{u,i}^k = l\middle|x_{u,i}^k = h\right)}_{p\left(y_{u,i}^k\middle|x_{u,i}^k = h\right)} \tag{10}$$

In Eq.(10), the likelihood $\lambda\left(y_{u,i}^k, y_{u',i'}^{k'}\right)$ is a similarity measure between the current and previous observations. It can be viewed as a (L+1)-D vector, each component of which corresponds to a specific value of $l$. Given $x_{u,i}^k = h$, the prior distribution of $z_{u,i}^k$ can also be viewed as a (L+1)-D vector. It is interesting to note that the likelihood $p\left(y_{u,i}^k\middle|x_{u,i}^k = h\right)$ is the inner product of the above two vectors, which is proportional to the cosine of the angle between them as the length of the vector $\lambda$ is constant with respect to $h$. By calculating this angular metric [32], the observation similarity measure $\lambda$ is transferred into likelihood scores for different label values, i.e., $p\left(y_{u,i}^k\middle|x_{u,i}^k = h\right)$.

3.3 Limiting the computational cost

From Eq.(1), we can see that the sampling space of labeling variable increases linearly with inference step. At step $k$, updating the belief state over sampling space costs $o(L)$ for likelihood evaluation, $o(kL)$ for prior and $o(k)$ for posterior. Here $L \leq k$ is the number of observations made on $u$'s neighbors up to step $k$, which is also increasing with time. The computational cost may become a limitation of the algorithm in large scale application. We overcome this problem by setting a limit $M$ of camera memory depth and a maximum size $H$ of the sampling space.

| |
|---|
| *Algorithm 1*: distributed inference for consistent labeling |
| 1:     For step $k = 1$ to $\infty$ (inference step index) |
| 2:         Cameras $u = 1$ to $N$ in parallel (camera index) |
| 3:             Await the event of object detection $y_{u,i}^k$; |
| 4:             Collect information from neighbors: observations $Y_{\mathcal{N}_u}^{k-1}$ and beliefs of corresponding labels; |
| 5:             Determine the sampling space of $x_{u,i}^k$ by (1); |
| 6:             Calculate the belief of $x_{u,i}^k$ by (10); |
| 7:         End parallel |
| 8:     End for. |

Fig.3 distributed inference algorithm

As older observations are less likely to be the immediate predecessor of the current one, for inference in step $k$, camera $u$ only collects the $M$ most recent observations and the corresponding belief states from its neighbors. Typically the number of objects is much less than the number of observations made in the networks. Thus limiting the size of sampling space by an appropriately chosen $H$ will not seriously affect the inference performance. At step $k$, we prune the sampling space $\Gamma(x_{u,i}^k)$ when its size exceeds $H$ by deleting the element with the lowest posterior probability. In this way, we obtain a distributed inference algorithm with constant computational and memory requirements. The complete algorithm is summarized in Fig.3.

We can see in Algorithm 1 that the inference is driven by the event of object detection. The algorithm advances one step further by inferring the label of the current observation. Moreover, the inference involves only local information processing on each camera node and communication between neighboring cameras. Thus, it is a complete distributed algorithm.

3.4 Missing detection

In the above discussion, we assume that objects can be detected reliably by smart cameras. In practice, however, false alarm and missing detection are always encountered due to unfavorable observing conditions. For consistent labeling, the problem of false alarm can be dealt with simply by deleting observations with likelihood below a specified threshold. On the other hand, the problem of missing detection is more critical and difficult to treat, as in this case the neighboring structure of the camera networks topology is destroyed, and the assumption underlying our distributed inference is violated. Thus, for the sake of briefness, we focus on missing detection only in this paper.

We partially overcome the problem of missing detection by considering information on the enlarged neighborhood of camera $u$ when the label of observation made on $u$ is inferred. We denote the enlarged neighborhood as $\mathcal{N}_u^q$, and call it as q-order neighborhood of $u$. $\mathcal{N}_u^q$ consists of all cameras $v$ in the networks that there exists at least one path between $u$ and $v$ with length no longer than $q$. The path length $q$ is defined as the number of camera nodes along a path between $u$ and $v$, and $q=0$ means $u$ and $v$ are connected directly. We assume that object cannot reach camera $u$ without being detected by cameras in $\mathcal{N}_u^q$, if it is not a newly appeared one. In this case, the spatio-temporal likelihood should be evaluated by the following mixture model

$$\lambda_{st}^q(d_{u,i}^k, d_{u',i'}^{k'}) = \omega_{uu'}^0 \lambda_{st}^0(d_{u,i}^k, d_{u',i'}^{k'}) + \sum_{r=1}^{q} \sum_{j} \omega_{uu'}^{r,j} \lambda_{st}^{r,j}(d_{u,i}^k, d_{u',i'}^{k'}) \tag{11}$$

the weight parameters $\omega$ correspond to different paths chosen by object when it is moving from $u'$ to $u$: $\omega_{uu'}^0$ is the probability of object moving along the edge $(u',u)$, which is the transition probability defined on the network topology. If $(u',u)$ does not exist, $\omega_{uu'}^0 = 0$. $\omega_{uu'}^{r,j}$ is the probability of object moving along the $j$ th path from $u'$ to $u$ with length $r$, which is the normalized product of the transition probabilities of the $r+1$ edges along that path. $\lambda_{st}^0$ is defined by Eq.(4), and $\lambda_{st}^{r,j}$ is the higher-order spatio-temporal observation model, which can be determined from the knowledge of monitored area, or directly from zero-order models. For example, if the traveling time along edges follows Gaussian, the traveling time along a $r$-length path also follows Gaussian, with mean and variance equaling to the sum of parameters corresponding to the $r+1$ edges in that path.

To apply the inference algorithm to the case of missing detection, we only need to replace the zero-order neighbors $\mathcal{N}_u$ with higher-order neighbors $\mathcal{N}_u^q$ in Algorithm 1, and evaluate the spatio-temporal likelihood by Eq.(11) instead of Eq.(4). After these modifications, the posterior of the labeling variable is inferred based on more information than before, leading to improved robustness against missing detection at the cost of increased communication and computation at each camera node.

## 4. Results

### 4.1 Experiment settings

In this section, we report the experimental results of the proposed algorithm in two different disjoint multi-camera surveillance scenarios. Object detection in single view is based on the background subtraction and shadow removal algorithm as proposed in [33]. Object tracking in each camera is based on the probabilistic, appearance based tracking algorithm as proposed in [34]. When a person is passing through the FOV of a camera, an observation is extracted from the collected video as described in Sec.2. Before conducting experiments, the observations extracted from about five hours-long video data collected by the camera networks are manually labeled and used for leaning CBTF and traveling time model $p(t_{u,i}^{en}|t_{u',i'}^{le})$ across neighboring cameras. The moving direction transition probabilities $p(e_{u,i}^{en}|e_{u',i'}^{le})$ are specified according to the layout of cameras in the monitored region.

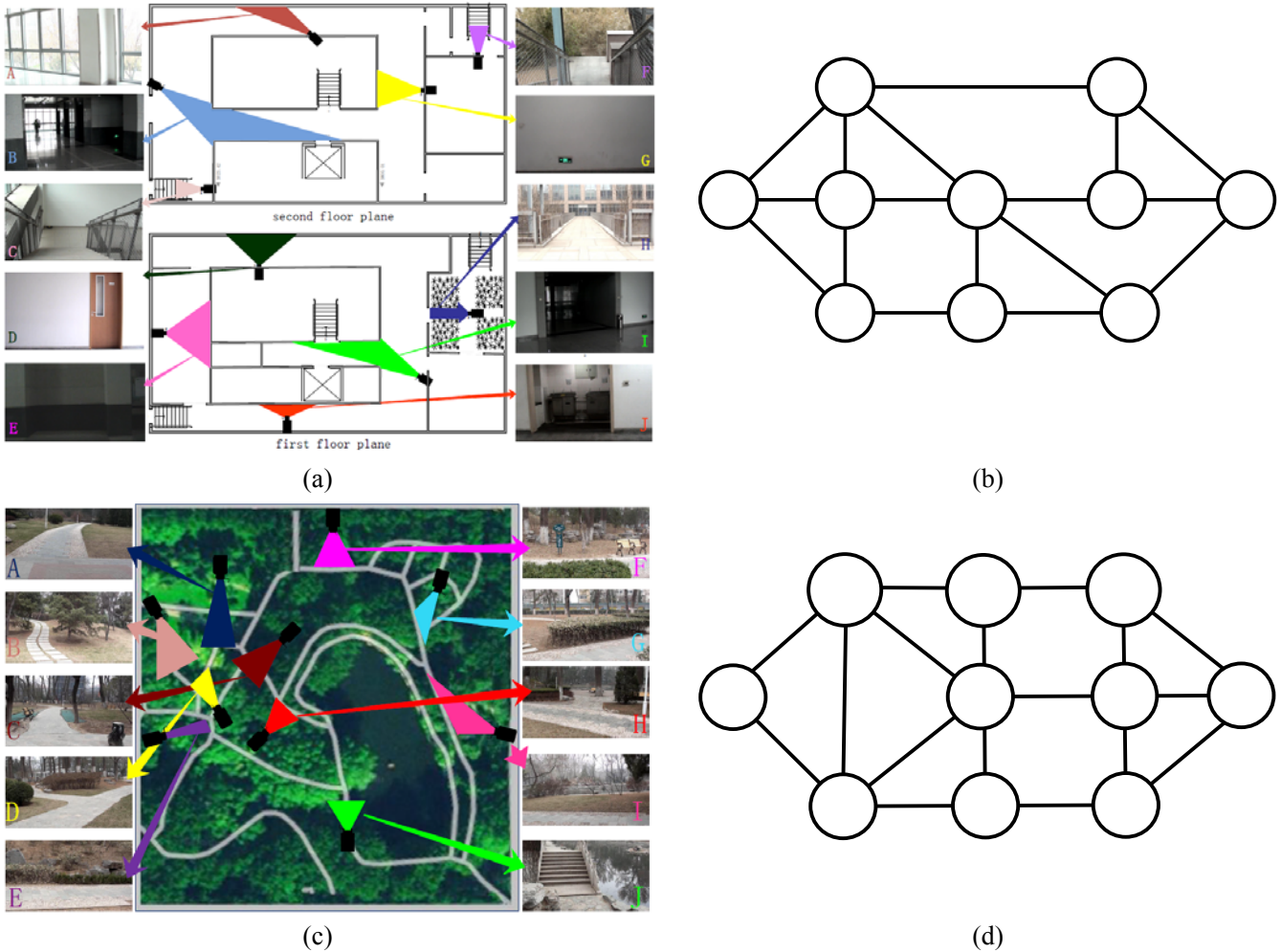

**Fig.4 Experiment settings. (a) Office building layout. (c) Campus garden layout. (b) and (d) Corresponding topology.**

*Scenario 1: Office Building experiment.* The experiment in scenario 1 was conducted with ten cameras mounted in an office building, five in the first floor and five in the second floor. The camera layout and corresponding topology are shown in Fig.4(a) and (b). 300 observations originated from 10 persons are extracted from the 70-minites video data collected by the cameras in the networks. The main challenges are the significant changes in illumination and view angles. For example, the areas covered by camera B and E are clearly dim due to the lack of lighting. And the view angles at stairs, C and F, are quite different form that at floor plane.

*Scenario 2: Campus Garden experiment.* The camera networks used in this experiment consists of ten cameras mounted in our campus garden, and the layout and corresponding topology are shown in Fig.4(c) and (d). During the experiment, 14 persons present in the monitored region. We gather altogether 300 observations of them from 90-minites video collected by the camera networks. In outdoor scenario, the illuminating conditions at each site are similar, but the monitored region is larger and the distance between cameras is longer than that of indoor case. Consequently, the variance of traveling time is large and the discrimination power of spatio-temporal cue is decreased.

4.2 Evaluation criteria

We use the following measures to evaluate the algorithms: the estimated number of objects $K$, the precision $P$, recall $R$, and F-measure of the reconstructed trajectories. Let $\{Y_1^*,\ldots,Y_{K^*}^*\}$ be the ground truth, each of which is a set of observations that belong to the same object. Let $\{Y_1,\ldots,Y_K\}$ denote the mutually disjoint subsets of $Y$, each of which is a set of observations assigned with the same label. The precision and recall of a set of estimated trajectories is defined as

$$P = \frac{1}{K}\sum_{i=1}^{K}\max_{j}\frac{|Y_i \cap Y_j^*|}{|Y_i|}, \qquad R = \frac{1}{K}\sum_{i=1}^{K}\max_{j}\frac{|Y_i \cap Y_j^*|}{|Y_j^*|}, \qquad F = \frac{2 \cdot P \cdot R}{P + R} \qquad (12)$$

The precision and recall represent the fidelity and the completeness of the estimated trajectories, respectively. A reconstructed trajectory with a single observation has a 100% precision, and a reconstructed trajectory with all observations has a 100% recall. And the F-measure is the harmonic mean of these two measures.

To evaluate the speed of the labeling algorithm, we measure its execution time $\tau(Y)$ on the observation set $Y$. The time cost involved with data collection, communication, object segmentation and tracking of object on single camera are ignored in this paper. For centralized algorithm, we use $\tau_c(Y)$, the execution time when the algorithm runs on a single machine, as speed measure. For distributed algorithm, which allows each camera agent to independently process data, we use $\tau_d(Y)$, the maximum of the execution time of individual agent, as speed measure.

4.3 Results

We apply our method to the two datasets. In Office Building experiment, we set the memory depth $M = 20$ and the sampling space limit $H = 15$. In Campus Garden experiment, we set them as $M = 25$ and $H = 20$, respectively. The resulting marginal distributions of labeling variables corresponding to each observation are shown in Fig.5. The observation's label is determined by choosing the label value with the maximum posterior marginal probability. The number of persons is determined by counting the number of different label values taken by all of the observations. The quantitative criteria are reported in Table 1 and Table 2. For both datasets, our method can estimates the number of persons correctly and gives quite satisfactory labeling results in terms of precision, recall and F-measure. Fig. 6 and Fig.7 show selected frames from the simultaneous video streams generated by the camera networks in the two experiments. Bounding boxes indicate detected objects and alphabetical labels indicate estimated identities.

Table 1 Results on Office building data ( $K^* = 10$ )

|  | $\tau_c$ (s) | $\tau_d$ (s) | K | Precision (%) | Recall (%) | F measure (%) |
|---|---|---|---|---|---|---|
| Zejdel [8] | 2.9x10$^5$ | X | 11 | 86.78 | 84.66 | 85.71 |
| Wan [31] | X | 10.89 | 13 | 75.96 | 69.00 | 72.31 |
| Ours | X | 5.82 | 10 | 92.84 | 92.00 | 92.42 |

Table 2 Results on Campus garden data ( $K^* = 14$ )

|  | $\tau_c$ (s) | $\tau_d$ (s) | K | Precision (%) | Recall (%) | F measure (%) |
|---|---|---|---|---|---|---|
| Zejdel [8] | 3.0 x10$^5$ | X | 16 | 95.44 | 97.62 | 96.52 |
| Wan [31] | X | 12.00 | 19 | 80.94 | 89.95 | 85.65 |
| Ours | X | 5.80 | 14 | 90.54 | 93.57 | 92.03 |

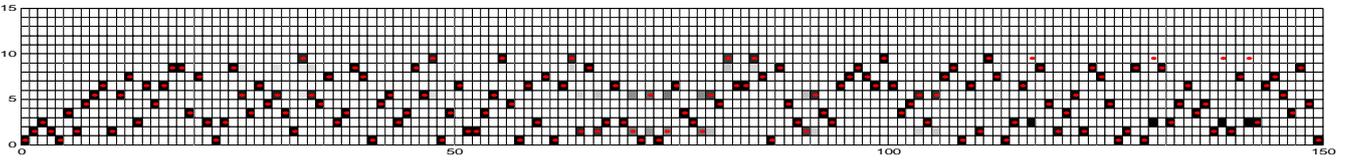
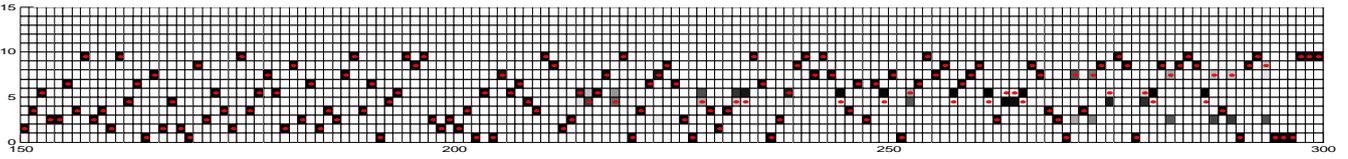

(a)

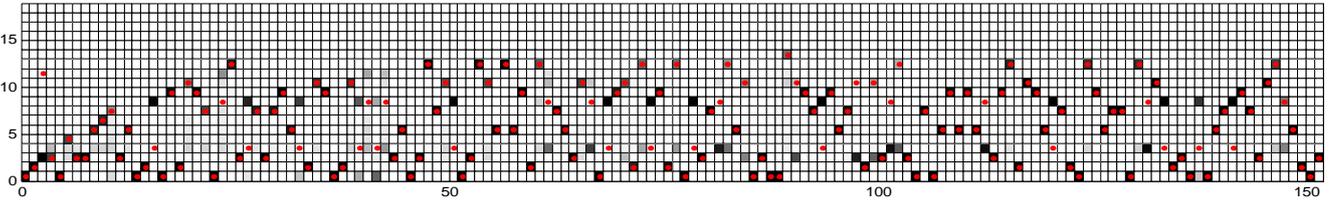
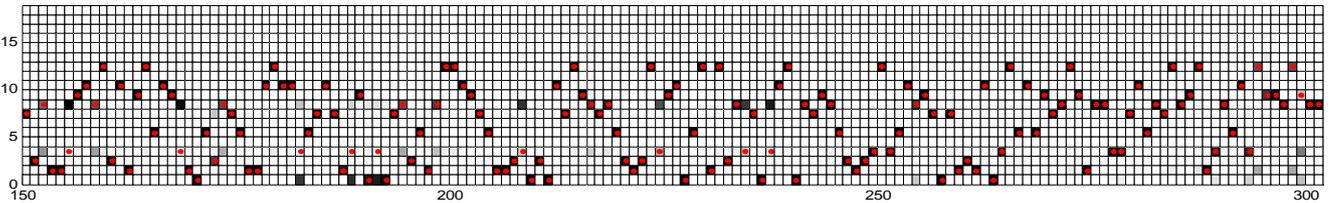

(b)

**Fig.5 Marginal distribution of labeling variable. Each column corresponds to one observation, sorted in time order. The true label of each observation is depicted by red star. Grayscale corresponds to the posterior probability of labeling variables. Black represent probability 1, and white 0. (a) Results on Office building dataset; (b) Results on Campus garden dataset.**

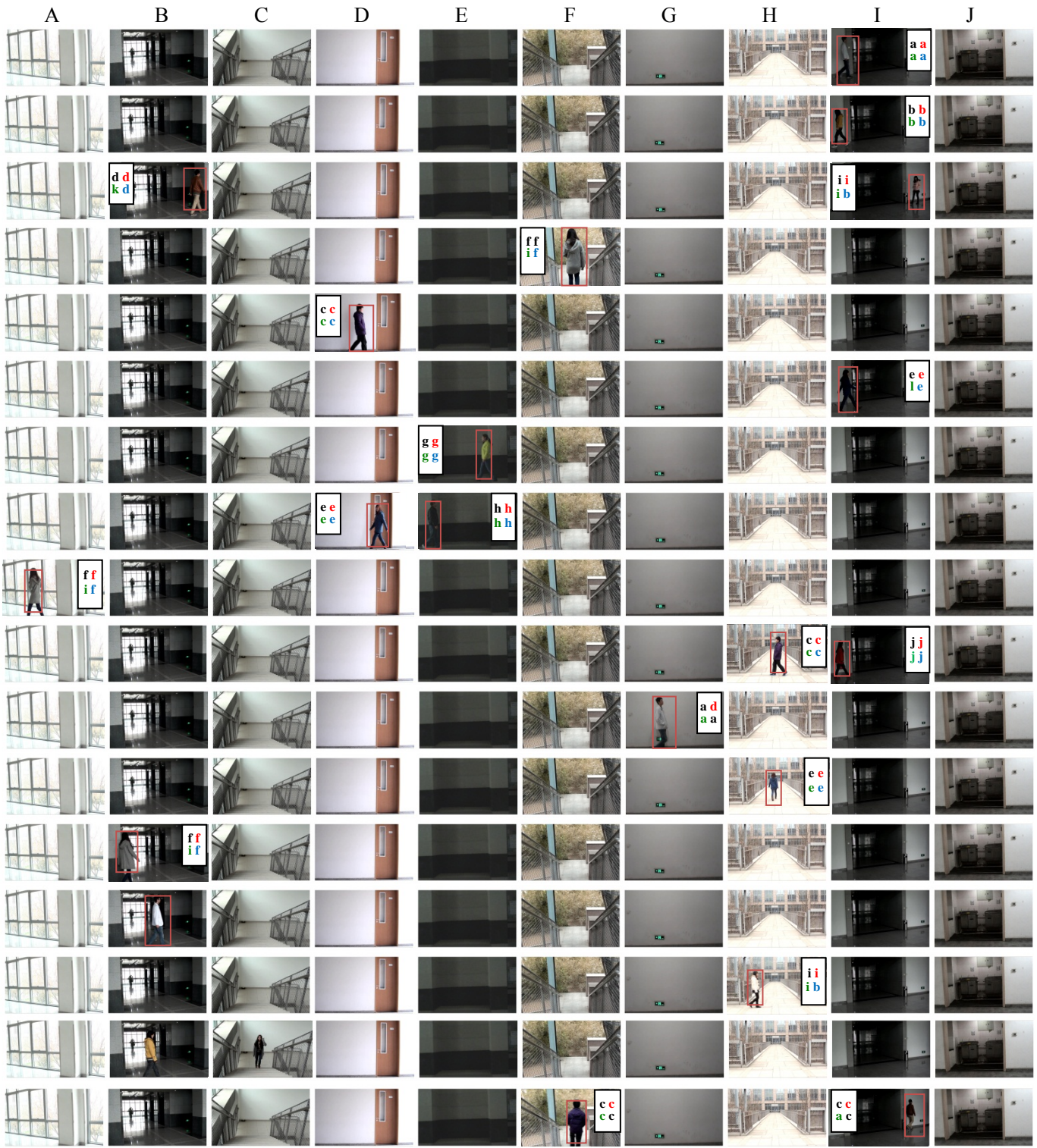

**Fig.6** Selected frames in Office Building experiment. Column corresponds to camera site, row corresponds to time instant. Detected person is shown with bounding box, the label of which is shown in the text box. Left-top, true label; right-top, result of Zejdel [8]; left-bottom, result of Wan [31]; right-bottom, our result.

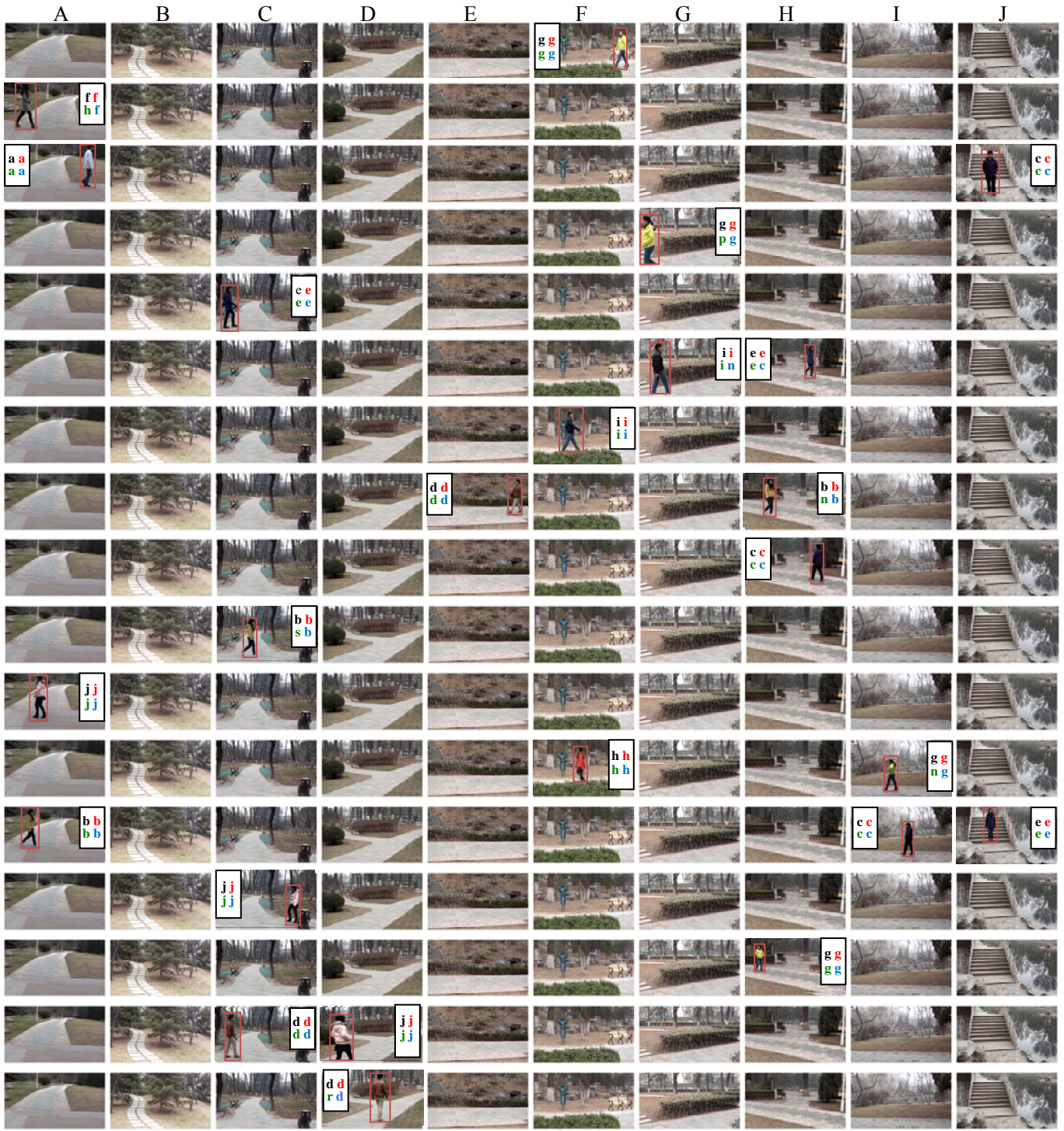

Fig.7 Selected frames in Campus Garden experiment. Column corresponds to camera site, row corresponds to time instant. Detected person is shown with bounding box, the label of which is shown in the text box. Left-top, true label; right-top, result of Zejdel [8]; left-bottom, result of Wan [31]; right-bottom, our result.

*Comparison with centralized inference algorithm* [8]. A closely related method to ours was proposed in [8], in which the joint distribution of observations and hidden variables is encoded by a dynamic Bayesian networks, and the posterior marginal of labeling and counter variables are inferred by using the Assumed Density Filter. By inferring the counter variables, the number of objects can be estimated from observations automatically. In [8], the appearance of a single person under different camera sites is assumed to follow a single Gaussian with Normal-Inverse Wishart distributed parameters (i.e. mean and covariance), allowing the appearance model to be updated analytically. However, the inference algorithm in [8] is centralized, and the computational and memory cost increase rapidly with number of data. To make it executable on our datasets, we limit the memory depth of the central server to 25 for both experiments, and set the maximum number of objects as 15 and 20 for the two experiments, respectively. It is obvious in Table 1 and 2 that the speed of [8] is much slower than ours, mainly due to its centralized nature. It is also noticeable that the centralized method [8] does not show superiority to our distributed method on the Office Building data in term of labeling accuracy, due to the unrealistic appearance assumption and truncating of history observations during inference. In fact, for centralized inference, a much deeper memory is required to ensure that the true predecessor of the current observation should not be discarded. But this will lead to unacceptable computational and memory cost.

*Comparison with enhanced distributed EM* [31]. We also compare our method with [31], in which a distributed inference algorithm is proposed, based on the same non-missing detection assumption as ours, to calculate the posterior distribution of labeling variables conditioned on both appearance and spatio-temporal observations. The appearance of a single person is also assumed to follow a single Gaussian, which is updated in M-step in a distributed EM framework. The method in [31] is offline, and requires the number of objects to be prefixed. In our experiments, we set the maximum object number as 15 and 20 respectively, run EM 30 iterations and initialize the appearance model by k-means clustering. As shown in Table 1 and 2, [31] cannot estimate the number of persons correctly in both experiments, and gives significantly lower labeling accuracy than our method. The inferior performance of [31] can be attributed to its single Gaussian assumption of appearance and its blindness in choosing the sampling space of labels. In experiments, we observe that by using [31] the trajectory of a single person tends to break into several pieces especially when his/her appearance undergoes obvious changes across camera sites, and consequently more trajectories are recovered than the ground truth. In contrast, our method can deal with this better due to the flexibility in appearance likelihood evaluation.

4.4 Missing detection

In our experiments, no missing detection occurs as the scenes are relatively clean. However, in practice, the interested objects may be miss-detected due to occlusions in crowd or low quality of the video. To verify our method in these cases, we randomly delete some of the observations, and apply the proposed algorithm (0-order model) and its modification (1-order model), respectively, to the remaining part of the two datasets. The average F-measure of 10 trials is shown in Fig.8. As expected, when missing detection occurs, the labeling accuracy of 0-order model based algorithm decreases rapidly. When 40 observations are deleted randomly (corresponding to a 13% missing rate), 0-order method gives poor results of 59% and 62% F-measure for the two datasets. In contrast, by enlarging the neighborhood, 1-order method can achieve 77% and 82% F-measure in this case. This can be attributed to the fact that in case of missing detection, the real predecessor of current observation is more likely to exist in higher-order neighborhood than in 0-order neighborhood, hence a correct link is more likely to be established by considering higher order neighborhood.

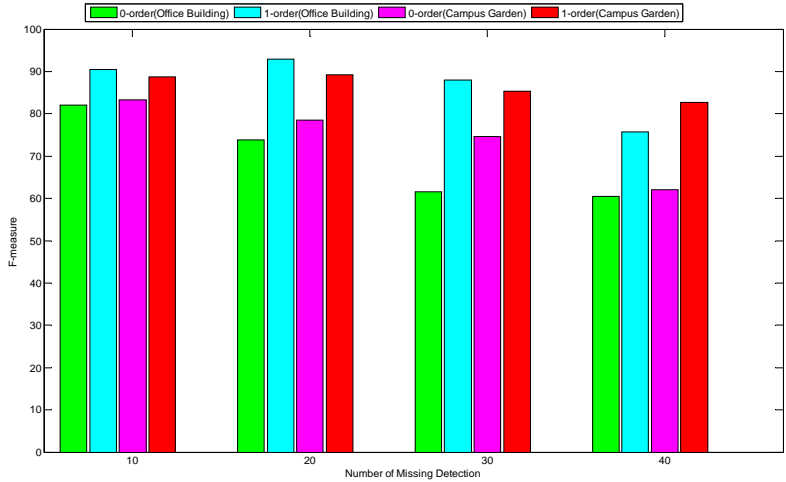

**Fig.8 F-measure in case of missing detection. X-axis: the number of missing detections. Legend: 0-order (Office Building) means result of our algorithm with 0-order model applied to Office Building data, and so on.**

5. Conclusion

In this paper, we present a distributed Bayesian inference framework for consistent labeling of tracked objects in non-overlapping camera networks. In this method, each camera in the networks performs inference on labeling variables over the online determined sampling space, based on local information and that collected from its neighbors. The similarity between pairs of observations is used for defining the appearance likelihood function in the framework, making it very flexible and particularly suitable in case of large observing condition variations across camera views. To cope with missing detection, we enlarge the neighborhood of each camera from which it collects information during inference, and use a higher order mixture model to evaluate spatio-temporal likelihood, leading to improved robustness of the algorithm. The effectiveness of the propose method is verified on two real datasets. In the future, we plan to investigate how to extend the use of our method to more realistic scenarios, in which the size of networks is larger, the duration of video collection is longer, and the camera scene is more crowded.

Acknowledgement

The authors are grateful to the student volunteers for their participation in the tracking experiments. This work is supported by the National Natural Science Foundation of China, under grant No.61174020.